# *Research Note*
# New Polynomial Classes for Logic-Based Abduction

**Bruno Zanuttini**                                                                ZANUTTI@INFO.UNICAEN.FR
*GREYC, Université de Caen, Boulevard du Maréchal Juin*
*14032 Caen Cédex, France*

## Abstract

We address the problem of propositional logic-based abduction, i.e., the problem of searching for a best explanation for a given propositional observation according to a given propositional knowledge base. We give a general algorithm, based on the notion of projection; then we study restrictions over the representations of the knowledge base and of the query, and find new polynomial classes of abduction problems.

## 1. Introduction

Abduction consists in searching for a plausible explanation for a given observation. For instance, if $p \models q$ then $p$ is a plausible explanation for the observation $q$. More generally, abduction is the process of searching for a set of facts (the *explanation*, here $p$) that, conjointly with a given *knowledge base* (here $p \rightarrow q$), imply a given *query* ($q$). This process is also constrained by a set of *hypotheses* among which the explanations have to be chosen, and by a preference criterion among them.

The problem of abduction proved its practical interest in many domains. For instance, it has been used to formalize text interpretation (Hobbs et al., 1993), system (Coste-Marquis & Marquis, 1998; Stumptner & Wotawa, 2001) or medical diagnosis (Bylander et al., 1989, Section 6). It is also closely related to configuration problems (Amilhastre et al., 2002), to the ATMS/CMS (Reiter & de Kleer, 1987), to default reasoning (Selman & Levesque, 1990) and even to induction (Goebel, 1997).

We are interested here in the complexity of *propositional logic-based* abduction, i.e., we assume both the knowledge base and the query are represented by propositional formulas. Even in this framework, many different formalizations have been proposed in the literature, mainly differing about the definition of an hypothesis and that of a best explanation (Eiter & Gottlob, 1995). We assume here that the hypotheses are the conjunctions of literals formed upon a distinguished subset of the variables involved, and that a best explanation is one no proper subconjunction of which is an explanation (*subset-minimality* criterion).

Our purpose is to exhibit new polynomial classes of abduction problems. We give a general algorithm for finding a best explanation in the framework defined above, independently from the syntactic form of the formulas representing the knowledge base and the query. Then we explore the syntactic forms that allow a polynomial running time for this algorithm. We find new polynomial classes of abduction problems, among which the one restricting the knowledge base to be given as a Horn DNF and the query as a positive CNF, and the one restricting the knowledge base to be given as an affine formula and the query as a disjunction of linear equations. Our algorithm also unifies several previous such results.

The note is organized as follows. We first recall the useful notions of propositional logic (Section 2), formalize the problem (Section 3) and briefly survey previous work about the complexity of abduction (Section 4). Then we give our algorithm (Section 5) and explore polynomial classes for it (Section 6). Finally, we discuss our results and perspectives (Section 7). For lack of space we cannot detail proofs, but a longer version of this work, containing detailed proofs and examples, is available (Zanuttini, 2003).

## 2. Preliminaries

We assume a countable number of propositional variables $x_1, x_2 \ldots$ and the standard connectives $\neg, \wedge, \vee, \oplus, \rightarrow, \leftrightarrow$. A *literal* is either a variable $x_i$ (*positive* literal) or its negation $\neg x_i$ (*negative* literal). A propositional formula is a well-formed formula built on a finite number of variables and on the connectives; $Var(\phi)$ denotes the set of variables that occur in the propositional formula $\phi$. A *clause* is a finite disjunction of literals, and a propositional formula is in *Conjunctive Normal Form (CNF)* if it is written as a finite conjunction of clauses. For instance, $\phi = (x_1 \vee \neg x_2) \wedge (\neg x_1 \vee x_2 \vee \neg x_3)$ is in CNF. The dual notions of clause and CNF are the notions of *term* (finite conjunction of literals) and *Disjunctive Normal Form (DNF)* (finite disjunction of terms).

An *assignment* to a set of variables $V$ is a set of literals $m$ that contains exactly one literal per variable in $V$, and a *model* of a propositional formula $\phi$ is an assignment $m$ to $Var(\phi)$ that satisfies $\phi$ in the usual way, where $m$ assigns 1 to $x_i$ iff $x_i \in m$; we also write $m$ as a tuple, e.g., 0010 for $\{\neg x_1, \neg x_2, x_3, \neg x_4\}$. We write $m[i]$ for the value assigned to $x_i$ by $m$, and $\mathcal{M}(\phi)$ for the set of all the models of a propositional formula $\phi$; $\phi$ is said to be *satisfiable* if $\mathcal{M}(\phi) \neq \emptyset$. A formula $\phi$ is said to *imply* a propositional formula $\phi'$ (written $\phi \models \phi'$) if $\mathcal{M}(\phi) \subseteq \mathcal{M}(\phi')$. More generally, we identify sets of models with Boolean functions, and use the notations $\overline{\mathcal{M}}$ (negation), $\mathcal{M} \vee \mathcal{M}'$ (disjunction) and so on.

The notion of *projection* is very important for the rest of the paper. For $m$ an assignment to a set of variables $V$ and $A \subseteq V$, write $Select_A(m)$ for the set of literals in $m$ that are formed upon $A$, e.g., $Select_{\{x_1,x_2\}}(0110) = 01$. Projecting a set of assignments onto a subset $A$ of its variables intuitively consists in replacing each assignment $m$ with $Select_A(m)$; for sake of simplicity however, we define the projection of a set of models $\mathcal{M}$ to be built upon the same set of variables as $\mathcal{M}$. This yields the following definition.

**Definition 1 (projection)** *Let $V = \{x_1, \ldots, x_n\}$ be a set of variables, $\mathcal{M}$ a set of assignments to $V$ and $A \subseteq V$. The* projection *of $\mathcal{M}$ onto $A$ is the set of assignments to $V$ $\mathcal{M}_{|A} = \{m \mid \exists m' \in \mathcal{M}, Select_A(m') = Select_A(m)\}$.*

For instance, let $\mathcal{M} = \{0001, 0010, 0111, 1100, 1101\}$ be a set of assignments to $V = \{x_1, x_2, x_3, x_4\}$, and let $A = \{x_1, x_2\}$. Then it is easily seen that

$$\mathcal{M}_{|A} = \{0000, 0001, 0010, 0011\} \cup \{0100, 0101, 0110, 0111\} \cup \{1100, 1101, 1110, 1111\}$$

since $\{Select_A(m) \mid m \in \mathcal{M}\} = \{00, 01, 11\}$.

Remark that the projection of the set of models of a formula $\phi$ onto a set of variables $A$ is the set of models of the most general consequence of $\phi$ that is independent of all the variables not in $A$. Note also that the projection of $\mathcal{M}(\phi)$ onto $A$ is the set of models of the formula obtained from $\phi$ by forgetting its variables not occurring in $A$. For more details about variable forgetting and independence we refer the reader to the work by Lang et al. (Lang et al., 2002).

It is useful to note some straightforward properties of projection. Let $\mathcal{M}, \mathcal{M}'$ denote two sets of assignments to the set of variables $V$, and let $A \subseteq V$. First, projection is distributive over disjunction, i.e., $(\mathcal{M} \vee \mathcal{M}')_{|A} = \mathcal{M}_{|A} \vee \mathcal{M}'_{|A}$. Now it is distributive over conjunction when $\mathcal{M}$ does not depend on the variables $\mathcal{M}'$ depends on, i.e., when there exist $A, A' \subseteq V$, $A \cap A' = \emptyset$ with $\mathcal{M}_{|A} = \mathcal{M}$ ($\mathcal{M}$ does not depend on $V \setminus A$) and $\mathcal{M}'_{|A'} = \mathcal{M}'$, $(\mathcal{M} \wedge \mathcal{M}')_{|A} = \mathcal{M}_{|A} \wedge \mathcal{M}'_{|A}$ holds; note that this is not true in the general case. Note finally that in general $(\overline{\mathcal{M}})_{|A}$ is not the same as $\overline{\mathcal{M}_{|A}}$.

## 3. Our Model of Abduction

We now formalize our model; for sake of simplicity, we first define *abduction problems* and then the notions of *hypothesis* and *explanation*.

**Definition 2 (abduction problem)** *A triple $\Pi = (\Sigma, \alpha, A)$ is called an* abduction problem *if $\Sigma$ and $\alpha$ are satisfiable propositional formulas and $A$ is a set of variables with $Var(\alpha), A \subseteq Var(\Sigma)$; $\Sigma$ is called the* knowledge base *of $\Pi$, $\alpha$ its* query *and $A$ its* set of abducibles.

**Definition 3 (hypothesis, explanation)** *Let $\Pi = (\Sigma, \alpha, A)$ be an abduction problem. An* hypothesis *for $\Pi$ is a set of literals formed upon $A$ (seen as their conjunction), and an hypothesis $E$ for $\Pi$ is an* explanation *for $\Pi$ if $\Sigma \wedge E$ is satisfiable and $\Sigma \wedge E \models \alpha$. If no proper subconjunction of $E$ is an explanation for $\Pi$, $E$ is called a* best explanation *for $\Pi$.*

Note that this framework does not allow one to specify that a variable must occur unnegated (resp. negated) in an explanation. We do not think this is a prohibiting restriction, since abducibles are intuitively meant to represent the variables whose values can be, e.g., modified, observed or repaired, and then no matter their sign in an explanation. But we note that it is a restriction, and that a more general framework can be defined where the abducibles are literals and the hypotheses, conjunctions of abducibles (Marquis, 2000).

We are interested in the computational complexity of computing a best explanation for a given abduction problem, or asserting there is none at all. Following the usual model, we establish complexities with respect to the size of the representations of $\Sigma$ and $\alpha$ and to the number of abducibles; for hardness results, the following associated decision problem is usually considered: is there at least one explanation for $\Pi$? Obviously, if this latter problem is hard, then the function problem also is.

## 4. Previous Work

The main general complexity results about propositional logic-based abduction with subset-minimality preference were stated by Eiter and Gottlob (1995). The authors show that deciding whether a given abduction problem has a solution at all is a $\Sigma_2^P$-complete problem, even if $A \cup Var(\alpha) = Var(\Sigma)$ and $\Sigma$ is in CNF. As stated as well by Selman and Levesque (1990), they also establish that this problem becomes "only" NP-complete when $\Sigma$ is Horn, and even acyclic Horn. Note that when SAT and deduction are polynomial with $\Sigma$ the problem is obviously in NP.

In fact, very few classes of abduction problems are known to be polynomial for the search for explanations. As far as we know, the only such classes are those defined by the following restrictions (once again we refer the reader to the references for definitions):

- $\Sigma$ is in 2CNF and $\alpha$ is in 2DNF (Marquis, 2000, Section 4.2)

- $\Sigma$ is given as a monotone CNF and $\alpha$ as a clause (Marquis, 2000, Section 4.2)

- $\Sigma$ is given as a definite Horn CNF and $\alpha$ as a conjunction of positive literals (Selman & Levesque, 1990; Eiter & Gottlob, 1995)

- $\Sigma$ is given as an acyclic Horn CNF with pseudo-completion unit-refutable and $\alpha$ is a variable (Eshghi, 1993)

- $\Sigma$ has bounded induced kernel width and $\alpha$ is given as a literal (del Val, 2000)

- $\Sigma$ is represented by its set of characteristics models (with respect to a particular basis) and $\alpha$ is a variable (Khardon & Roth, 1996); note that a set of characteristic models is *not* a propositional formula, but that the result is however similar to the other ones

- $\Sigma$ is represented by the set of its models, or, equivalently, by a DNF with every variable occurring in each term, and $\alpha$ is any propositional formula.

The first two classes are proved polynomial with a general method for solving abduction problems with the notion of prime implicants, the last one is obvious since all the information is explicitely given in the input, and the four others are exhibited with *ad hoc* algorithms.

Let us also mention that Amilhastre et al. (2002) study most of the related problems in the more general framework of multivalued theories instead of propositional formulas, i.e., when the domain of the variables is not restricted to be $\{0, 1\}$. The authors mainly show, as far as this note is concerned, that deciding whether there exists an explanation is still a $\Sigma_2^P$-complete problem (Amilhastre et al., 2002, Table 1).

Note that not all these results are stated in our exact framework in the papers cited above, but that they all still hold in it. Let us also mention that the problem of *enumerating* all the best explanations for a given abduction problem is of great interest; Eiter and Makino (2002) provide a discussion and some first results about it, mainly in the case when the knowledge base is Horn.

## 5. A General Algorithm

We now give the principle of our algorithm. Let us stress first that, as well as, e.g., Marquis' construction (Marquis, 2000, Section 4.2), its outline matches point by point the definition of a best explanation; our ideas and Marquis' are anyway rather close.

We are first interested in the hypotheses in which every abducible $x \in A$ occurs (either negated or unnegated); let us call them *full hypotheses*. Note indeed that every explanation $E$ for an abduction problem is a subconjunction of a full explanation $F$; indeed, since $E$ is by definition such that $\Sigma \wedge E$ is satisfiable and implies $\alpha$, it suffices to let $F$ be $Select_A(m)$ for a model $m$ of $\Sigma \wedge E \wedge \alpha$. Minimization of $F$ will be discussed later on.

**Proposition 1** *Let $\Pi = (\Sigma, \alpha, A)$ be an abduction problem, and $F$ a full hypothesis of $\Pi$. Then $F$ is an explanation for $\Pi$ if and only if there exists an assignment $m$ to $Var(\Sigma)$ with $F = Select_A(m)$ and $m \in \mathcal{M}(\Sigma) \wedge \overline{(\mathcal{M}(\Sigma \wedge \overline{\alpha}))_{|A}}$.*

**Proof** Assume first $F$ is an explanation for $\Pi$. Then (i) there exists an assignment $m$ to $Var(\Sigma)$ with $m \models \Sigma \wedge F$, thus $F = Select_A(m)$ and $m \in \mathcal{M}(\Sigma)$, and (ii) $\Sigma \wedge F \models \alpha$, i.e., $\Sigma \wedge F \wedge \overline{\alpha}$ is unsatisfiable, thus $F \notin \{Select_A(m) \mid m \in \mathcal{M}(\Sigma \wedge \overline{\alpha})\}$, thus $m \notin (\mathcal{M}(\Sigma \wedge \overline{\alpha}))_{|A}$, thus $m \in \overline{(\mathcal{M}(\Sigma \wedge \overline{\alpha}))_{|A}}$. Conversely, if $m \in \mathcal{M}(\Sigma) \wedge \overline{(\mathcal{M}(\Sigma \wedge \overline{\alpha}))_{|A}}$ let $F = Select_A(m)$. Then we have (i) since $m \in \mathcal{M}(\Sigma)$, $\Sigma \wedge F$ is satisfiable, and (ii) since $m \notin (\mathcal{M}(\Sigma \wedge \overline{\alpha}))_{|A}$, there is no $m' \in \mathcal{M}(\Sigma \wedge \overline{\alpha})$ with $Select_A(m') = F$, thus $\Sigma \wedge F \wedge \overline{\alpha}$ is unsatisfiable, thus $\Sigma \wedge F \models \alpha$. □

Thus we have characterized the full explanations for a given abduction problem. Now minimizing such an explanation $F$ is not a problem, since the following greedy procedure, given by Selman and Levesque (1990) reduces $F$ into a best explanation for $\Pi$:

   **For** every literal $\ell \in F$ **do**
     **If** $\Sigma \wedge F \backslash \{\ell\} \models \alpha$ **then** $F \leftarrow F \backslash \{\ell\}$ **endif**;
   **Endfor**;

Note that depending on the order in which the literals $\ell \in F$ are considered the result may be different, but that in all cases it will be a best explanation for $\Pi$.

Finally, we can give our general algorithm for computing a best explanation for a given abduction problem $\Pi = (\Sigma, \alpha, A)$; its correctness follows directly from Proposition 1:

   $\Sigma' \leftarrow$ a propositional formula with $\mathcal{M}(\Sigma') = \mathcal{M}(\Sigma) \wedge \overline{(\mathcal{M}(\Sigma \wedge \overline{\alpha}))_{|A}}$;
   **If** $\Sigma'$ is unsatisfiable **then return** "No explanation";
   **Else**
     $m \leftarrow$ a model of $\Sigma'$;
     $F \leftarrow Select_A(m)$;
     minimize $F$;
     **return** $F$;
   **Endif**;

## 6. Polynomial Classes

We now explore the new polynomial classes of abduction problems that our algorithm allows to exhibit. Throughout the section, $n$ denotes the number of variables in $Var(\Sigma)$.

### 6.1 Affine Formulas

A propositional formula is said to be *affine* (or in *XOR-CNF*) (Schaefer, 1978; Kavvadias & Sideri, 1998; Zanuttini, 2002) if it is written as a finite conjunction of linear equations over the two-element field, e.g., $\phi = (x_1 \oplus x_3 = 1) \wedge (x_1 \oplus x_2 \oplus x_4 = 0)$. As can be seen, equations play the same role in affine formulas as clauses do in CNFs; roughly, affine formulas represent conjunctions of parity or equivalence constraints. This class proves interesting for knowledge representation, since on one hand it is tractable for most of the common reasoning tasks, and on the other hand the affine approximations of a knowledge base can be made very small and are efficiently learnable (Zanuttini, 2002). We show that projecting an affine formula onto a subset of its variables is quite easy too, enabling our algorithm to run in polynomial time. The proof of the following lemma is easily obtained with gaussian elimination (Curtis, 1984): triangulate $\phi$ with the variables in $A$ put rightmost, and then keep only those equations formed upon $A$; full details are given in the technical report version (Zanuttini, 2003).

**Lemma 1** *Let $\phi$ be an affine formula containing $k$ equations, and $A \subseteq Var(\phi)$. Then an affine formula $\psi$ with $\mathcal{M}(\psi) = (\mathcal{M}(\phi))_{|A}$ and containing at most $k$ equations can be computed in time $O(k^2|Var(\phi)|)$.*

**Proposition 2** *If $\Sigma$ is represented by an affine formula containing $k$ equations and $\alpha$ by a disjunction of $k'$ linear equations, and $A$ is a subset of $Var(\Sigma)$, then searching for a best explanation for $\Pi = (\Sigma, \alpha, A)$ can be done in time $O((k+k')((k+1)^2 + |A|(k+k'))n)$.*

**Sketch of proof** It is easily seen that an affine formula (containing $k' + k$ equations and $n$ variables) for $\Sigma \wedge \overline{\alpha}$ can be computed in time linear in the size of $\alpha$; this formula can be projected onto $A$ in time $O((k+k')^2 n)$, and we straightforwardly get a disjunction of at most $k + k'$ linear equations for $\overline{(\mathcal{M}(\Sigma \wedge \overline{\alpha}))_{|A}}$. Then we can use distributivity of $\wedge$ over $\vee$ for solving the satisfiability problem of the algorithm; recall that SAT can be solved in time $O(k^2 n)$ for an affine formula of $k$ equations over $n$ variables by the elimination method of Gauss (Curtis, 1984). The remaining operations are straightforward. □

Note that variables, literals and clauses are special cases of disjunctions of linear equations.

### 6.2 DNFs

Though the class of DNF formulas has very good computational properties, abduction remains a hard problem for it as a whole, even with additional restrictions. Recall that the TAUTOLOGY problem is the one of deciding whether a given DNF formula represents the identically true function, and that this problem is coNP-complete.

**Proposition 3** *Deciding whether there is at least one explanation for a given abduction problem $(\Sigma, \alpha, A)$ is NP-complete when $\Sigma$ is given in DNF, even if $\alpha$ is a variable and $A \cup \{\alpha\} = Var(\Sigma)$.*

**Sketch of proof** Membership in NP is obvious, since deduction with DNFs is polynomial; now it is easily seen that $\Sigma$ is tautological if and only if the abduction problem $(\Sigma \vee (x), x, Var(\Sigma))$ has no explanation, where $x$ is a variable not occuring in $\Sigma$ (see the DNF $\Sigma \vee (x)$ as the implication $\overline{\Sigma} \to x$); $\Sigma \vee (x)$ is in DNF, and we get the result. □

However, when $\Sigma$ is represented by a DNF projecting it onto $A$ is easy; indeed, the properties of projection show that it suffices to cancel its literals that are not formed upon $A$. Consequently, if $\phi$ is such a DNF containing $k$ terms, then a DNF $\psi$ with $\mathcal{M}(\psi) = (\mathcal{M}(\phi))_{|A}$ and containing at most $k$ terms can be computed in time $O(k|Var(\phi)|)$.

Thus we can show that some subclasses of the class of all DNFs allow polynomial abduction. We state the first result quite generally, but note that its assumptions are satisfied by natural classes of DNFs: e.g., that of *Horn* DNFs, i.e., those DNFs with at most one positive literal per term; similarly, that of Horn-renamable DNFs, i.e., those that can be turned into a Horn DNF by replacing some variables with their negation, and simplifying double negations, everywhere in the formula; 2DNFs, those DNFs with at most two literals per term. We omit the proof of the following proposition, since it is essentially the same as that of Proposition 2 (simply follow the execution of the algorithm).

**Proposition 4** *Let $\mathcal{D}$ be a class of DNFs that is stable under removal of occurrences of literals and for which the TAUTOLOGY problem is polynomial. If $\Sigma$ is restricted to belong to $\mathcal{D}$, $\alpha$ is a clause and $A$ is a subset of $Var(\Sigma)$, then searching for a best explanation for $\Pi = (\Sigma, \alpha, A)$ can be done in polynomial time.*

Thus we can establish that abduction is tractable if (among others) $\Sigma$ is in Horn-renamable DNF (including the Horn and reverse Horn cases) or in 2DNF, and $\alpha$ is a clause.

Finally, let us point out that with a very similar proof we can obtain polynomiality for some problems obtained by strengthening the restriction of Proposition 4 over $\Sigma$, but weakening that over $\alpha$.

**Proposition 5** *If $\Sigma$ is represented by a Horn (resp. reverse Horn) DNF of $k$ terms and $\alpha$ by a positive (resp. negative) CNF of $k'$ clauses, and $A$ is a subset of $Var(\Sigma)$, then searching for a best explanation for $\Pi = (\Sigma, \alpha, A)$ can be done in time $O((k+|A|)kk'n)$. The same holds if $\Sigma$ is represented by a positive (resp. negative) DNF of $k$ terms and $\alpha$ by a Horn (resp. reverse Horn) CNF of $k'$ clauses.*

Once again note that variables, literals and terms are all special cases of (reverse) Horn CNFs, and that variables, positive (resp. negative) clauses and positive (resp. negative) terms are all special cases of positive (resp. negative) CNFs.

## 7. Discussion and Perspectives

The general algorithm presented in this note allows us to derive new polynomial restrictions of abduction problems; even if this is not discussed here, for lack of space, it also allows to unify some previously known such restrictions (such as $\Sigma$ in 2CNF and $\alpha$ in 2DNF, or $\Sigma$ in monotone CNF and $\alpha$ given as a clause). The following list summarizes the main new polynomial restrictions:

- $\Sigma$ given as an affine formula and $\alpha$ as a disjunction of linear equations (Proposition 2)
- $\Sigma$ in Horn-renamable DNF and $\alpha$ given as a clause (Proposition 4)
- $\Sigma$ in 2DNF and $\alpha$ given as a clause (Proposition 4)
- $\Sigma$ in Horn (reverse Horn) DNF and $\alpha$ in positive (negative) CNF (Proposition 5)
- $\Sigma$ in negative (positive) DNF and $\alpha$ in reverse Horn (Horn) CNF (Proposition 5).

Moreover, even if there is no guarantee for efficiency in the general case the presentation of our algorithm does not depend on the syntactic form of $\Sigma$ or $\alpha$, and it uses only standard operations on Boolean functions (projection, conjunction, negation).

Another interesting feature of this algorithm is that before minimization it computes the explanations *intentionnally*. Consequently, all the full explanations can be enumerated with roughly the same delay as the models of the formula representing them ($\Sigma'$). However, of course, there is no guarantee that two of them would not be minimized into the same *best* explanation, which prevents from concluding that our algorithm can enumerate all the *best* explanations; trying to extend it into this direction would be an interesting problem. For more details about enumeration we refer the reader to Eiter and Makino's work (Eiter & Makino, 2002).

As identified by Selman and Levesque (1990), central to the task is the notion of projection over a set of variables, and our algorithm isolates this subtask. However, our notion of projection only concerns variables, and not literals, which prevents from imposing a sign to the literals the hypotheses are formed upon, contrariwise to more general formalizations proposed for abduction, as Marquis' (Marquis, 2000). Even if we think this is not a prohibiting restriction, it would be interesting to try to fix that weakness of our algorithm while preserving its polynomial classes.

Another problem of interest is the behaviour of our algorithm when $\Sigma$ and $\alpha$ are not only propositional formulas, but more generally *multivalued theories*, in which the domain of variables is not restricted to be $\{0, 1\}$: e.g., signed formulas (Beckert et al., 1999). This framework is used, for instance, for configuration problems by Amilhastre et al. (2002). It is easily seen that our algorithm is still correct in this framework; however, there is still left to study in which cases its running time is polynomial.

Finally, problems of great interest are those of deciding the *relevance* or the *necessity* of an abducible (Eiter & Gottlob, 1995). An abducible $x$ is said to be *relevant* to an abduction problem $\Pi$ if there is at least one best explanation for $\Pi$ containing $x$ or $\neg x$, and *necessary* to $\Pi$ if all the best explanations for $\Pi$ contain $x$ or $\neg x$. It is easily seen that $x$ is necessary for $\Pi = (\Sigma, \alpha, A)$ if and only if $\Pi' = (\Sigma, \alpha, A \backslash \{x\})$ has no explanation, hence showing that polynomial restrictions for the search for explanations are polynomial as well for deciding the necessity of an hypothesis as soon as they are stable under the substitution of $A \backslash \{x\}$ for $A$, which is the case for all restrictions considered in this note. Contrastingly, we do not know of any such relation for relevance, and the study of this problem would also be of great interest.

## Acknowledgments

The author wishes to thank the anonymous referees of this version and those of a previous one (Proc. JNPC'02, in French), as well as Jean-Jacques Hébrard, for very valuable and constructive comments.